\documentclass[10pt,twocolumn,letterpaper]{article}

\usepackage{iccv}              

\usepackage{algorithm}  
\usepackage{algpseudocode}

\usepackage{pifont}
\newcommand{\cmark}{\ding{51}} 
\newcommand{\xmark}{\ding{55}} 
\usepackage{multirow}
\usepackage{xcolor}

\definecolor{iccvblue}{rgb}{0.21,0.49,0.74}
\usepackage[pagebackref,breaklinks,colorlinks,allcolors=iccvblue]{hyperref}

\def\confName{ICCV}
\def\confYear{2025}

\title{All-in-One Transfer Image Compression from Human Perception to Multi-Machine Perception}

\author{Jiancheng Zhao \quad Xiang Ji \quad Yinqiang Zheng\thanks{Corresponding author. Email: yqzheng@ai.u-tokyo.ac.jp}\\
The University of Tokyo, Tokyo, Japan\\
}

\begin{document}
\maketitle

\begin{abstract}
Efficiently transferring Learned Image Compression (LIC) model from human perception to machine perception is an emerging challenge in vision-centric representation learning.
Existing approaches typically adapt LIC to downstream tasks in a single-task manner, which is inefficient, lacks task interaction, and results in multiple task-specific bitstreams. 
In this paper, we propose a \emph{multi-task adaptation framework} that enables transferring a pre-trained base codec to multiple machine vision tasks through a unified model and a single training process.
To achieve this, we design an \emph{asymmetric adaptation architecture} consisting of a task-agnostic encoder adaptation and task-specific decoder adaptation. 
Furthermore, we introduce two feature propagation modules to facilitate \emph{inter-task} and \emph{inter-scale} feature represenation learning. 
Experiments on \emph{PASCAL-Context} and \emph{NYUD-V2} dataset demonstrate that our method outperforms both Fully Fine-Tuned and other Parameter Efficient Fine-Tuned (PEFT) baselines. Code will be released.
\end{abstract}

\section{Introduction}
\label{sec:intro}

In recent years, Image Coding for Machine (ICM) has seen a surge in demand, driven by the rapid advancement of deep learning technologies. As a result, an increasing volume of visual data is being processed by machines instead of being viewed by humans.
The primary goal of ICM is to learn compact visual representations that minimize transmission cost while preserving high accuracy for downstream vision tasks.
To this end, a straightforward solution is to customize the encoder, decoder, or bitstream in a task-specific manner~\cite{chamain2021endtoendoptimizedimagecompression,bai2021endtoendimagecompressionanalysis,Choi_2022,feng2022imagecodingmachinesomnipotent,feng2025semanticallystructuredimagecompression,liu2023icmh,liu2021semantics,Fischer_2024}.
\begin{table}[t]
\centering

\label{tab:method_comparison}
\renewcommand{\arraystretch}{1.1}
\small
\begin{tabular}{@{}p{0.55\linewidth}cc@{}}
\toprule
\textbf{Method} & \textbf{Efficient} & \textbf{MT Transfer} \\
\midrule
Full-Finetune                                              & \xmark & \xmark \\
ICMH-Net~\cite{liu2023icmh}                                            & \cmark & \xmark \\
TransTIC~\cite{chen2023transtictransferringtransformerbasedimage}    & \cmark & \xmark \\
AdaptICMH~\cite{li2024imagecompressionmachinehuman}             & \cmark & \xmark \\

\textbf{Ours}                                                                  & \cmark & \cmark \\
\bottomrule
\end{tabular}
\caption{Comparison of representative codec transfer methods in terms of parameter efficiency and multi-task transferability. A method is considered \textit{Efficient} if it enables adaptation with only a small number of trainable parameters, and supports \textit{Multi-Task (MT) Transfer} if it allows joint optimization across multiple downstream tasks.}
\end{table}
However, such task-specific customization often results in costly training procedures, increased bitrate consumption, and significant parameter overhead. 
To address these limitations, recent studies have turned to a more efficient alternative. Motivated by the observation that widely available base codecs—originally optimized for human perception—can already produce high-quality reconstructions that preserve most of the semantic content in images, these works explore parameter-efficient fine-tuning (PEFT) as a means to adapt such codecs directly for machine vision tasks. 
By updating only a small number of parameters, PEFT-based methods aim to retain the original human-perception quality of the base codec, while efficiently transferring the latent features and the reconstructed images into a more semantically compact space that is better aligned with downstream machine vision tasks. 
Despite the efficiency advantages of PEFT-based adaptation, existing methods predominantly focus on single-task transfer, treating each task in a completely independent and isolated manner, as summarized in~\Cref{tab:method_comparison}, which incurs several limitations:
(1) Each task still requires separately customized training, which is inefficient and hinders scalability as the number of tasks grows;
(2) The lack of cross-task interaction prevents the model from leveraging shared representations across related task domains;
(3) Task-specific adaptations produce separate bitstreams, leading to increased storage requirements and greater complexity in deployment.

To address these limitations, we propose a \textbf{multi-task adaptation} framework that enables transferring a pre-trained base codec to multiple vision tasks using a unified model and a single-stage training process. 
At the core of our design is a structurally decoupled architecture that introduces parallel, task-specific adaptors to preserve task specialization. Specifically, we redesign each decoding stage of the base codec as a parallel multi-path structure, where each path corresponds to a specific task and leverages an independent adaptor to extract task-specific features.
The extracted set of task features is subsequently processed through both inter-task and inter-scale information propagation. First, we introduce a \emph{Task Aggregation Module}, which integrates features from different tasks and applies a channel selection mechanism to allow each task to access and incorporate complementary information from others. Second, we introduce a \emph{Multi-Scale Fusion} Module, which applies cross-scale attention to fuse and selectively refine features across different scales. This enables the integration of information from diverse receptive fields.

Overall, our contributions are as follows:
\begin{itemize}
    \item A \emph{multi-task adaptation framework} that enables transferring a pre-trained base codec to multiple vision tasks within a single model and a single-stage training process.
    \item Two modules to enhance representation learning: \emph{Task Aggregation Module} and \emph{Multi-Scale Fusion}, for cross-task and cross-scale interaction, respectively.
    \item We validate our approach on two widely-used multi-task learning benchmarks, \emph{PASCAL-Context} and \emph{NYUD-V2}, demonstrating superior performance.
\end{itemize}

\section{Related Works}
\label{sec:rela}

\paragraph{\textbf{Learned Image Compression (LIC)}}
Learned Image Compression (LIC) was first introduced by~\citet{ballé2017endtoendoptimizedimagecompression}, which adopts an autoencoder-based architecture to perform transform coding in the pixel space. Due to its superior rate-distortion (R-D) performance, LIC has shown great potential as a promising alternative to traditional image compression paradigms. A typical LIC codec consists of three key components: an analysis transform that maps the input image from the high-dimensional pixel space to a compact latent representation; an entropy model that encodes the latent variables into a compressed bitstream; and a synthesis transform that reconstructs the image from the latent space back to the pixel domain. 
The main research directions in LIC can be broadly categorized into two types. The first focuses on designing more efficient and expressive codec architectures, including more representative analysis and synthesis transforms. These structures have evolved from early CNN-based~\cite{ballé2017endtoendoptimizedimagecompression,ballé2018variationalimagecompressionscale,cui2022asymmetricgaineddeepimage,mentzer2020highfidelitygenerativeimagecompression,cheng2020learnedimagecompressiondiscretized} designs to Transformer-based~\cite{lu2021transformerbasedimagecompression,li2024frequencyawaretransformerlearnedimage,zou2022devildetailswindowbasedattention,liu2023learnedimagecompressionmixed} models, enabling better modeling capacity. In addition, recent efforts explore user-controllable compression, such as variable-rate coding~\cite{yang2022slimmablecompressiveautoencoderspractical,choi2019variable,li2024onceforallcontrollablegenerativeimage} and distortion–perception~\cite{agustsson2023multirealismimagecompressionconditional} trade-off control, to enhance flexibility in practical applications. 
The second line of research focuses on designing more powerful entropy models to better estimate the probability distribution of latent representations. This has evolved from factorized~\cite{ballé2017endtoendoptimizedimagecompression} and hyperprior-based~\cite{ballé2018variationalimagecompressionscale} models to more advanced autoregressive entropy models~\cite{lee2019contextadaptiveentropymodelendtoend,he2021checkerboardcontextmodelefficient,Jiang_2023,qian2022entroformertransformerbasedentropymodel,minnen2020channelwiseautoregressiveentropymodels,minnen2018jointautoregressivehierarchicalpriors}.
owever, most LIC methods are human-centric, as they are typically optimized using perceptual quality metrics such as MSE or LPIPS. While effective for human viewing, such objectives may not align with the needs of machine vision. In particular, pixel-wise distortion metrics tend to over-allocate bits to visually fine-grained details, while potentially neglecting semantically important structures that are critical for downstream vision tasks

\paragraph{\textbf{Image Compression for Machine Vision}}
The rapid growth of computer vision applications has led to an increasing portion of visual data being consumed by machines rather than humans—for example, in autonomous driving, traffic monitoring, and visual surveillance. This shift motivates the need for image compression systems optimized for machine vision, aiming to jointly minimize transmission cost while preserving task-relevant information for downstream analysis.
A straightforward solution is to define distinct encoder–decoder pairs tailored to each specific task. These approaches can be further categorized into two types: (1) Independent designs~\cite{liu2022improving,chamain2020endtoendoptimizedimagecompression,Fischer_2024,chamain2021endtoendoptimizedimagecompression}, where each task is handled by a separately trained codec; and
(2) Scalable designs~\cite{yan2021sssic,feng2025semanticallystructuredimagecompression,bai2021endtoendimagecompressionanalysis}, which share part of the model parameters while adapting to different tasks via task-specific components.
However, this line of work often leads to multiple task-specific bitstreams and requires dedicated design for each task, resulting in increased system complexity and poor scalability as the number of tasks grows.
An alternative line of work aims to extract a unified bitstream for multiple tasks. Specifically, these methods design a task-general encoder to produce semantically compact and generalizable representations that serve as a common source for all tasks. Task-specific outputs are then generated by applying either task-specific decoders~\cite{feng2022imagecodingmachinesomnipotent,liu2024ratedistortioncognitioncontrollableversatileneural} or task-guided transformations~\cite{Choi_2022} to adapt the shared representation to the needs of each downstream task.
However, such codecs designed specifically for multiple vision tasks typically require non-trivial codec architecture design and computationally expensive training, making it difficult to scale to many tasks and challenging to deploy in practice.

\paragraph{\textbf{Transfer Human Perception to Machine Peception}}
Most learned image compression (LIC) models are optimized for human perception, typically using distortion metrics such as MSE or LPIPS. Recent works have made progress by adopting tuning frameworks that transfer a pre-trained human-centric codec to various machine vision tasks. This pipeline enables ICMH (Image Compression for both Human and Machine) by training only a lightweight task-specific module while keeping the base codec frozen for shared use across human and machine perception. Unfortunately, these methods only support single-task optimization, requiring customized training for each task and producing separate bitstreams, which is not storage-efficient and leads to a linearly increasing training burden as the number of tasks scales up.

\begin{figure*}[t]
  \includegraphics[width=\textwidth,height=0.37\textheight]{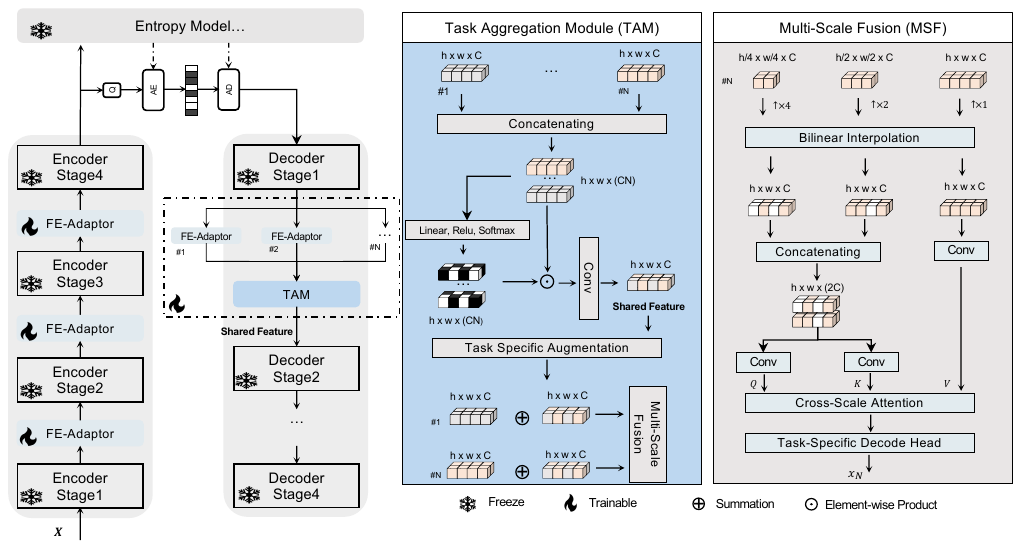}
  \caption{\textbf{Overview of our proposed architecture.} The backbone adopts a standard autoencoder design based on~\emph{Lu2022-TIC}, consisting of an encoder and a decoder. The encoder follows a single-path structure, with a task-agnostic \emph{Feature Extraction Adaptor} inserted after each encoding stage. In contrast, the decoder follows a multi-path design, where each path contains a task-specific \emph{Feature Extraction Adaptor}. The resulting task features are further processed by the \emph{Task Aggregation Module} and the \emph{Multi-Scale Fusion Module} to enable both inter-task and inter-scale information propagation.}
  \label{fig:overview}
\end{figure*}

\section{Method}
\label{sec:method}


\paragraph{\textbf{Problem Formulation}}
Given a pre-trained base codec optimized for human perception, our goal is to efficiently transfer it to multiple machine vision tasks within a single unified model and through a single-stage training process. 
This poses a fundamental challenge: the model must strike a delicate balance between task feature \emph{sharing} and \emph{decoupling}. 
On the one hand, it should leverage the overlapping visual cues across vision tasks; on the other hand, it must avoid excessive coupling, which can lead to negative task interference.
To achieve this, we adopt a multi-task adaptation framework that promotes feature sharing across tasks while preserving task-specific representations through structural decoupling.
Furthermore, the training of our framework follows the parameter-efficient fine-tuning (PEFT) paradigm by introducing lightweight plug-in modules, enabling effective adaptation without compromising the original rate-distortion performance of the base codec.

\paragraph{\textbf{Method Overview}} 
We begin by reviewing the architecture of the base codec and its training objective, which is optimized for human perception.
We then present an overview of our proposed multi-task adaptation framework.
Subsequently, we detail the design of the adaptation components, including the \emph{Task Aggregation Module} and the \emph{Multi-Scale Fusion Module}, which facilitate inter-task and inter-scale feature interaction.
Finally, we describe the joint optimization strategy used to train our model.

\subsection{Preliminary} 
\label{method:preliminary}
Learned Image Compression (LIC) frameworks typically adopt an autoencoder-based architecture, composed of an analysis network $g_a$ and a synthesis network $g_s$. The encoder $g_a$ maps the input image $x$ from the high-dimensional pixel space to a more compact latent representation $y$, i.e., $y = g_a(x)$. The decoder $g_s$ then reconstructs the image from the quantized latent $\hat{y}$, i.e., $\hat{x} = g_s(\hat{y})$.
Naively storing the quantized latent $\hat{y}$ can incur significant storage overhead. To address this, LIC models the distribution of symbols in $\hat{y}$ via a learned entropy model $p(\hat{y})$, enabling efficient entropy coding. Throughout the compression process, the goal is to achieve a balance between compression rate and reconstruction quality. This is formulated as an end-to-end optimization of the Rate-Distortion (R-D) trade-off:

\begin{equation}
    \mathcal{L}_{rd} = \mathbb{E}_{x \sim p_x}[r(\hat{y}) + \lambda\cdot\mathcal{D}(x,\hat{x})]
    \label{equ:rd}
\end{equation}

where $r(\hat{y})$ denotes the estimated bitrate, $\mathcal{D}(x, \hat{x})$ is a distortion metric (e.g., MSE or LPIPS), and the hyperparameter $\lambda$ controls the trade-off between compression rate and reconstruction quality.

In this work, we adopt~\emph{Lu2022-TIC}~\cite{lu2021transformerbasedimagecompression} as our base codec, which integrates convolutional layers and Transformer blocks in a hybrid architecture. This design enables the model to capture both local and long-range dependencies, supporting effective multi-scale feature extraction. Each encode and decode stage leverages a Residual Swin Transformer Block (RSTB) after a convolutional transformation, formulated as:
\begin{equation}
    \mathbf{F}_i = \text{RSTB}({\text{Conv}(\mathbf{F}_{i-1})})
\end{equation}
where $\mathbf{F}_{i-1}$ is the input feature from the previous stage, $\text{Conv($\cdot$)}$ adjusts the spatial resolution, and $\text{RSTB}$ captures contextual information through shifted window-based self-attention.

\subsection{Framework Overview}
\label{method:overview}
As shown in~\Cref{fig:overview}, given a pre-trained image codec optimized for human perception (e.g., using MSE or LPIPS as the distortion metric), our goal is to adapt it efficiently to multiple downstream vision tasks within a single model. 

To achieve this, we propose an asymmetric adaptation framework that extends the frozen base codec with two targeted architectural modifications.
In the encoding stage, our goal is to enable generalizable feature extraction. To this end, we insert a single lightweight and task-agnostic \textbf{Feature Extraction adaptor} into each encoding stage (we deliberately avoid task-specific modulation to prevent multiple bitstreams).
In the decoding stage, to support task-specific specialization, we insert multiple \emph{parallel} Feature Extraction adaptors to independently extract task-specific features for each task. These features are then passed to the \textbf{Task Aggregation Module}, where information is exchanged across tasks to formulate a shared representation and enhance each task’s feature by incorporating complementary cues from other tasks.
Finally, the shared feature is propagated as the primary representation to the subsequent decoding stage.
Task-specific features produced at different stages are organized into a multi-scale hierarchy, which is further processed by a \textbf{Multi-Scale Fusion} Module. The fused task representations are then forwarded to task-specific decode heads to generate the final predictions for each task.


\subsection{Task Aggregation Module}
\label{method:adaptor}
\paragraph{\textbf{Feature Extraction Adaptor}}
Feature Extraction (FE) adaptor is inserted into each encoding stage and is also placed in parallel along each task-specific path in the decoding stage. Within our framework, it serves as a basic unit for extracting informative representations. 
In this paper, we directly adopt the FE adaptor design proposed~\cite{li2024imagecompressionmachinehuman} as our FE adaptor, which employs a dual-branch architecture to simultaneously capture spatial and frequency-domain features. Given an intermediate feature map \(\mathbf{F}^{\text{in}} \in \mathbb{R}^{H \times W \times C}\), the adaptor processes it through two parallel streams:
\begin{align}
\mathbf{F}^{\text{out}} &=
\underbrace{\text{DWConv}(\mathbf{F}^{\text{in}})}_{\text{Spatial Branch}} 
+ 
\underbrace{\text{IFFT}(\mathbf{M} \odot \text{FFT}(\mathbf{F}^{\text{in}}))}_{\text{Frequency Branch}}
\end{align}

Here, FFT and IFFT denote the Fast Fourier Transform and its inverse, respectively. \(\text{DWConv}(\cdot)\) represents depth-wise convolution, and \(\odot\) indicates element-wise multiplication. \(\mathbf{M}_i\) is a learnable spectral mask that adaptively emphasizes or suppresses specific frequency components.

\paragraph{\textbf{Task Aggregation Module (TAM)}}

Each decoding stage consists of $N$ parallel Feature Extraction (FE) adaptors, each responsible for generating task-specific features for the corresponding task. To enable cross-task information propagation, these features are passed into a Task Aggregation Module (TAM), where inter-task relationships and dependencies are captured. Within the TAM, task-specific features are further refined by selectively incorporating relevant information from other tasks.
In particular, the TAM recieve as input $N$ task-specific features, each of shape $h \times w \times C$, denoted as $\{F_{k}\}_{k=1}^{N}$. These features are first concatenated along the channel dimension to form a unified representationand $\mathbf{F}_{\text{cat}} \in \mathbb{R}^{h \times w \times (NC)} $. $\mathbf{F}_{\text{cat}}$ is then passed through a transformation module $\Phi(\cdot) $, which consists of linear and nonlinear layers followed by a softmax activation, producing a channel-wise attention mask of the same shape. The attention mask is used to reweight the concatenated features via element-wise multiplication, resulting in a shared feature representation:
\begin{align}
\mathbf{F}_{\text{cat}} &= \text{Concat}(\mathbf{F}_1, \ldots, \mathbf{F}_N) \\
\mathbf{F}_{\text{shared}} &= \mathbf{F}_{\text{cat}} \odot 
\text{Softmax} \left( \Phi(\mathbf{F}_{\text{cat}}) \right)
\end{align}
The resulting \emph{shared feature} follows two propagation paths. First, it is propagated through the main path of the subsequent decoding stages. Second, it simultaneously serves as residual information to inject cross-task knowledge into each task-specific representation. Specifically, the shared feature is directly added to each task-specific feature to extract complementary information from other tasks and enhance the original representations. The refined features are then passed to the multi-scale fusion module, which will be described in the following section.





\subsection{Multi-Scale Fusion}
While the TAM within each decoding stage enables task interactions, it remains limited to a certain scale. However, prior work~\cite{vandenhende2020mtinetmultiscaletaskinteraction} has shown that task affinity patterns are scale-dependent, which means the mutual influence between relevant tasks is not necessarily preserved across different scales. Thus, we propose a \textbf{Multi-Scale Fusion (MSF)} module that explicitly incorporates cross-scale information propagation.
As illustrated in~\Cref{fig:overview} (right), for individual task, the MSF module takes as input a set of multi-scale features $\{\mathbf{F}_{s}\}_{s=1}^{3}$, obtained from different decoding stages.
We treat features with large and medium receptive fields, $\mathbf{F}_1$ and $\mathbf{F}_2$ as the query and key, and the feature with a the smallest receptive field, $\mathbf{F}_3$, as value, to formulate the query-key-value pair.
To achieve this, we first align the spatial resolutions of $\mathbf{F}_1$, $\mathbf{F}_2$, $\mathbf{F}_3$, using bilinear interpolation. Then, we apply three separate convolutional layers to extract the query and key from the concatenated features of $\mathbf{F}_1$ and $\mathbf{F}_2$, and the value from $\mathbf{F}_3$. Finally, a cross-scale attention module is employed to produce representations that integrate information across different receptive field scales. 
Formally, this process is defined as:
\begin{equation}
\mathbf{Q}, \mathbf{K} = \text{Conv}_{q,k}([\mathbf{F}_1; \mathbf{F}_2]), \quad \mathbf{V} = \text{Conv}_{v}(\mathbf{F}_3)
\end{equation}
\begin{equation}
\mathbf{F}_{\text{out}} = \text{Softmax}\left( \frac{\mathbf{Q} \cdot \mathbf{K}^\top}{\sqrt{d}} \right) \cdot \mathbf{V}
\end{equation}

Here, $\text{Conv}_{q,k}(\cdot)$ and $\text{Conv}_v(\cdot)$ are $1 \times 1$ convolutions for query/key and value projection, $[\cdot;\cdot]$ denotes channel-wise concatenation, and $d$ is the channel dimension used for scaling.




\subsection{Optimization Strategy}
\label{method:optimization}

Our framework is trained end-to-end to jointly optimize compression efficiency and downstream task performance. To quantify the compression efficiency, we adopt the rate-distortion formulation defined in~\Cref{equ:rd}, which balances the trade-off between bitrate and reconstruction quality.
To incorporate supervision from downstream tasks, we define the task loss as a weighted combination of task-specific objectives:
\begin{equation}
    \mathcal{L}_{\text{task}} = \sum_{i=1}^{T} w_i \cdot \mathcal{L}_i,
\end{equation}
where \(\mathcal{L}_i\) is the loss for the \(i\)-th task and \(w_i\) controls its relative contribution.

The overall training objective combines both components: 
\begin{equation}
    \mathcal{L}_{\text{total}} = \lambda_{\text{rd}} \cdot \mathcal{L}_{rd} +  
    \mathcal{L}_{\text{task}},
    \label{equ:loss}
\end{equation}
where \(\lambda_{\text{rd}}\) balances the trade-off between compression and task performance. This formulation encourages the model to learn compact, bit-efficient representations that remain informative and effective for multiple downstream tasks.

\section{Experiments}
\label{sec:exp}

\begin{figure*}[t]
  \includegraphics[width=\textwidth]{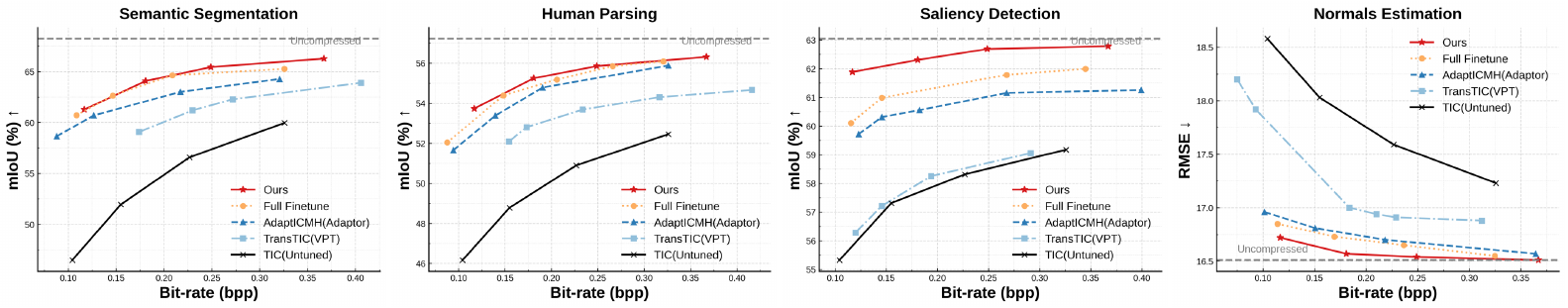}
  \caption{Rate–Accuracy performance comparison across four machine vision tasks on PASCAL-Context.
}
  \label{fig:pascal_rd}
\end{figure*}

\begin{table*}[htp]
\centering

\label{tab:pascal_bd}

\resizebox{1\textwidth}{!}{
\begin{tabular}{l|rr|rr|rr|rr|r|c}
\hline
\multirow{2}{*}{\textbf{Method}} & \multicolumn{2}{c|}{\textbf{Semantic Segmentation}} & \multicolumn{2}{c|}{\textbf{Human Parsing}} & \multicolumn{2}{c|}{\textbf{Saliency Detection}} & \multicolumn{2}{c|}{\textbf{Normals Estimation}} & \textbf{Trainable$\downarrow$} & \textbf{Multi-Task} \\
& BD-Rate$\downarrow$ & BD-Acc$\uparrow$ & BD-Rate$\downarrow$ & BD-Acc$\uparrow$ & BD-Rate$\downarrow$ & BD-Acc$\uparrow$ & BD-Rate$\downarrow$ & BD-Acc$\uparrow$ & Params (M) & /$\Delta$m(\%) \\
\hline
Full Fine-Tuning 
&-74.63\% &9.63   &-70.76\% &5.23   &-72.12\% &3.53 &- &1.09     &30.04(100\%) & \xmark \\

TransTIC  
&-39.99\% &4.36   &-49.16\% &2.74 &-6.00\% &0.23    &-56.47\% &0.73     & 6.48(21.4\%)  &\xmark \\

AdaptICMH    &-67.39\% &8.45   &-63.82\%  &4.78   &-77.62\% &2.76 &-77.91\% &1.09     &1.15(3.7\%) & \xmark \\

\hline

\textbf{Ours}  
&-74.89\% &9.61  &-78.91\% &5.29   &- &4.53 &-  &1.20  & 0.63 (2.1\%) & +0.25 \\
\hline

\end{tabular}}
\caption{Comparison of Rate–Accuracy performance and trainable parameters across four tasks on PASCAL-Context. BD-Rate and BD-Acc are calculated with the base codec as anchor. Arrows indicate the preferred direction ($\downarrow$: lower is better, $\uparrow$: higher is better). (–) indicates unavailableBD-Rate. $\Delta$m represents the BD-Acc improvement of our multi-task model over the single-task baseline (AdaptICMH). \xmark\ indicates that multi-task learning is not supported.
}
\end{table*}

\begin{figure*}[t]
  \includegraphics[width=\textwidth]{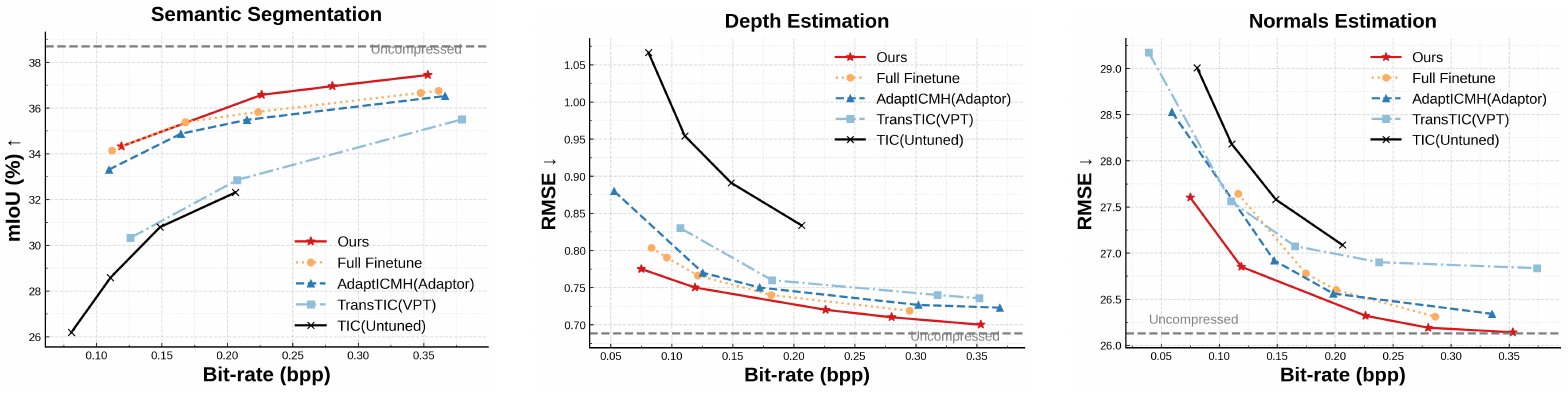}
  \caption{Rate–Accuracy performance comparison across different machine vision tasks on NYUD-V2.
}
  \label{fig:nyud_rd}
\end{figure*}

\begin{table*}[htp]
\centering

\label{tab:nyud_bd}

\resizebox{1\textwidth}{!}{
\begin{tabular}{l|rr|rr|rr|r|c}
\hline
\multirow{2}{*}{\textbf{Method}} & \multicolumn{2}{c|}{\textbf{Semantic Segmentation}} & \multicolumn{2}{c|}{\textbf{Depth Estimation}} & \multicolumn{2}{c|}{\textbf{Normals Estimation}} & \textbf{Trainable$\downarrow$} & \textbf{Multi-Task} \\
& BD-Rate$\downarrow$ & BD-Acc$\uparrow$ & BD-Rate$\downarrow$ & BD-Acc$\uparrow$ & BD-Rate$\downarrow$ & BD-Acc$\uparrow$ & Params (M) & /$\Delta$m(\%) \\
\hline
Full Fine-Tuning 
&-67.45\% &3.61   &-69.10\% &0.16   &-21.80\% &0.51     &30.04(100\%) & \xmark \\

TransTIC
&-10.82\% &0.57   &-45.00\% &0.11   &-25.79\% &0.52     & 4.86(16.18\%)  &\xmark \\

AdaptICMH     
&-60.42\% &3.24   &-58.97\% &0.16   &18.50\% &0.76     &0.86(2.9\%) & \xmark \\

\hline

\textbf{Ours}  
&-66.06\% &4.06   &-79.67\% &0.18   &-46.70\%  &1.11  & 0.50 (1.7\%) & +0.28 \\
\hline
\end{tabular}}
\caption{Comparison of Rate–Accuracy performance and trainable parameters across three tasks on NYUD-V2.}
\end{table*}

\subsection{Experimental Setup}
\paragraph{\textbf{Dataset}} 
We conduct our experimental evaluation on two widely used benchmarks in multi-task learning: PASCAL-Context~\cite{everingham2010pascal} and NYUD-V2~\cite{Silberman:ECCV12}.
On PASCAL-Context, we evaluate four tasks: \emph{Semantic Segmentation, Human Parsing, Surface Normal Estimation}, and \emph{Saliency Detection}. 
On NYUD-V2, we consider three tasks: \emph{semantic segmentation, depth estimation, Surface Normals Estimation}.
To meet codec constraints—which typically require image dimensions divisible by 64—all images are resized to \(512 \times 512\) during both training and testing.

\paragraph{\textbf{Evaluation Metrics}} 
We use mean Intersection over Union (mIoU) to evaluate semantic segmentation, human parsing, and saliency detection. For depth estimation and surface normal estimation, we report Root Mean Square Error (RMSE) of the predicted depth and angle differences.  
In addition, to assess the overall effectiveness of our multi-task learning framework, we adopt the multi-task performance metric \(\Delta m\) introduced in~\cite{maninis2019attentivesingletaskingmultipletasks}, which quantifies the average performance gain over the corresponding single-task baselines. A higher \(\Delta m\) indicates better multi-task generalization. 
\vspace{-2mm}
\begin{equation}
    \Delta m = \frac{1}{T}\sum_{i=1}^{T}(-1)^{l_i}(M_i - M_{st,i})/M_{st,i}
\end{equation}
\vspace{-2mm}

where $l_i=1$ if lower is better for task $i$, and $l_i=0$ otherwise. Here, $M_i$ and $M_{st,i}$ denote the performance of the multi-task and single-task models, respectively. In our experiments, we use BD-Acc as the task performance metric $M_i$, AdaptICMH the single-task baseline.

\subsection{Implementation Details}
The training of \textbf{Our framework} strictly follows the coding-for-machine paradigm. Our method outputs a decoded image for each task (with one single inference), which is then processed by the downstream model. The loss is computed between downstream model’s predictions and its ground-truth labels, and the final training objective is a weighted sum of task-specific losses, with weights adopted from~\cite{vandenhende2020mtinetmultiscaletaskinteraction}. 
Gradient flow follows the parameter-efficient fine-tuning (PEFT) pipeline: gradients are propagated from the downstream models back through the compression model, including both the base codec and our tuning modules. 
Importantly, thanks to our parallel adaptor design, the gradients of each task-specific adaptor are primarily attributed to its corresponding task-specific loss, which equips the model with the capacity for task decoupling. Additionally, to benchmark our multi-task adaptation approach, we compare it with the following single-task tuning baselines:
(1) \textbf{Full Fine-tuning}: All model parameters are fine-tuned independently for each task on top of the pre-trained base codec. 
(2) \textbf{Adaptor Tuning}~\cite{li2024imagecompressionmachinehuman}: Several trainable lightweight adaptor modules are inserted into each encoder and decoder stage of the base codec.
(3) \textbf{Visual Prompt Tuning (VPT)}~\cite{chen2023transtictransferringtransformerbasedimage}: Trainable task-specific prompts are inserted into the transformer layers of the base codec.

All methods, including ours, are built upon the same base codec, \emph{Lu2022-TIC}, and initialized with the pre-trained weights provided in~\cite{chen2023transtictransferringtransformerbasedimage} to ensure a fair comparison.
For the Adaptor Tuning and VPT baselines, we follow their official implementations and insert the PEFT modules into the encoder and decoder stages according to the configurations described in their respective papers.

\paragraph{\textbf{Training Setup}}
All experiments are conducted using PyTorch on a single NVIDIA RTX 4090 GPU. Models are trained with the Adam optimizer, starting from an initial learning rate of $1 \times 10^{-4}$, and employing a multi-step learning rate decay at epochs 2, 4, and 8 with a decay factor of 0.1. The total number of training epochs is set to 10. We adopt the loss formulation described in~\Cref{equ:loss}, which combines the rate-distortion loss with task-specific objectives. 
To explore performance under different compression rates, we vary the rate-distortion trade-off coefficient $\lambda_{\text{rd}}$ in the range of 0.1 to 5.0.

\paragraph{\textbf{Downstream Task Models}}
For downstream task modeling, we adopted~\emph{MTLoRA}~\cite{agiza2024mtloralowrankadaptationapproach}, which utilizes a Swin-Tiny~\cite{liu2021swintransformerhierarchicalvision} backbone coupled with a task-specific decoder head adapted from HRNet~\cite{wang2020deephighresolutionrepresentationlearning}. The model is pre-trained on ImageNet-1K~\cite{deng2009imagenet} and then fine-tuned on multiple downstream tasks using task-specific LoRA~\cite{hu2021loralowrankadaptationlarge} modules.

\subsection{Main Results}
\label{sec:exp_main}
We compare our method against Full Fine-tuning, Adaptor Tuning, VPT, and the base codec (denoted as \emph{TIC (Untuned)}).
\Cref{fig:pascal_rd} and \Cref{fig:nyud_rd} show the Rate–Accuracy curves on PASCAL-Context and NYUD-V2, respectively, illustrating the performance of each baseline under varying compression rates.
Following prior work~\cite{chen2023transtictransferringtransformerbasedimage}, we additionally report BD-Rate and BD-Accuracy in~\Cref{tab:pascal_bd} and \Cref{tab:nyud_bd} to provide a quantitative comparison.
BD-Rate measures the average bitrate savings at equivalent accuracy, while BD-Accuracy reflects the average accuracy improvement at equivalent bitrate.
Additional qualitative results are provided in the appendix to further illustrate the visual performance of our method.

We highlight the following experimental results:

(1) Across all four/three downstream tasks and under varying compression rates, our method consistently outperforms all baselines. This demonstrates the strong generalization and task adaptability of our method.

(2) Our approach is rather more parameter-efficient, equiring only 2.1\%/1.7\% of the trainable parameters of the base codec—just half the size of the adaptor-based baseline.

(3) our approach improve storage efficiency and deployment scalability. Unlike baselines that require training and storing a separate bitstream for each task, our method produces a single shared bitstream that serves all tasks.

(4) Our architecture demonstrates beneficial task interactions. Our multi-task model achieves better performance than its single-task baseline, consistently across tasks and bitrates, as also reflected in the positive $\Delta m$ score.

\subsection{Ablation Study}
\label{sec:exp_abla}

\paragraph{\textbf{Task Number Scaling}}
To further investigate the effect of task interaction and demonstrate the robustness of our method under task scaling scenarios, we evaluate its performance with varying numbers of downstream tasks.
Specifically, we use semantic segmentation as the anchor task and incrementally add additional tasks to form four combinations: (1) +Human Parsing, (2) +Saliency Detection, (3) +Human Parsing + Saliency Detection, (4) +Human Parsing + Saliency Detection + Surface Normals.

We report both BD-Rate and BD-Acc for semantic segmentation and human parsing relative to the base codec, as summarized in~\Cref{tab:task_number}.
The results demonstrate that both tasks consistently benefit from multi-task interaction, with improved performance as more tasks are introduced.
For a more intuitive comparison, we also provide the corresponding rate–accuracy curves in the supplementary material.

\begin{table}[t]
\centering
\begin{tabular}{c|c|c}
\toprule
\textbf{TASK (SemSeg)} & \textbf{SemSeg} & \textbf{Human Parsing} \\
\midrule
\textit{+Human} &-69.82\%/9.36  &-71.01\%/5.04  \\
\textit{+Sal} &-64.00\%/9.33  &  --  \\
\textit{+Human/Sal} &-70.13\%/9.39  &-74.85\%/5.22  \\
\textit{+Human/Sal/Normal} & -74.89\% /9.61  & -78.91\%/5.29  \\
\bottomrule
\end{tabular}
\caption{BD-Rate and BD-Acc Performance of Semantic Segmentation and Human Parsing comparison under varing task number (set SemSeg as anchor task).}
\label{tab:task_number}
\end{table}

\paragraph{\textbf{Computation Complexity}}
Although our method requires the fewest trainable parameters to adapt the base codec, the parallel multi-path design of the Task Adaptors introduces additional architectural complexity, which may affect computational efficiency. To assess this impact, we follow prior works~\cite{chen2023transtictransferringtransformerbasedimage} and compare the complexity of competing methods using kilo multiply–accumulate operations per pixel (kMACs/pixel).

As shown in~\Cref{tab:complexity}, inserting the single-path S-Adaptor into the encoder introduces the smallest increase in complexity among all PEFT baselines (from 142.5 to 149.7 kMACs/pixel).
On the decoder side, the multi-path T-Adaptor introduces a higher increase in complexity, but within an acceptable range.
It is also worth noting that~\Cref{tab:complexity} compares our multi-task setup against baselines in single-task settings. Overall, our method still demonstrates better computational efficiency than other competing approaches.

\begin{table}[htp]
\centering

\begin{tabular}{c|cc|c}
\toprule
\textbf{Method} & \textbf{Encoder} & \textbf{Decoder} & \textbf{Params (M)} \\
\midrule
\textit{Full-Finetune} & 142.5 & 188.5 & 7.51 $\times$ 4 \\
\textit{TransTIC} & 332.0 & 202.6 & 1.62 $\times$ 4\\
\textit{AdaptICMH} & 157.2 & 203.2 & 0.29 $\times$ 4 \\

\hline
\textbf{Ours} & 149.7 & 236.1 & 0.63 $\times$ 1 \\
\bottomrule
\end{tabular}
\caption{Comparison of the kMACs/pixel and model size.}
\label{tab:complexity}
\end{table}

\paragraph{\textbf{Sparse Prediction Task}}
Owing to the reliance of our setting on existing multi-task learning benchmarks, most evaluations are limited to dense prediction tasks.
While the tasks we included could already reflect the model's ability to capture both global and local information—semantic segmentation, for instance, depends on a global receptive field (e.g., locating a person in the scene), whereas depth and surface normal estimation rely more on local receptive fields to capture fine-grained edge and texture details—we are further interested in understanding how well our framework generalizes to structurally different tasks.

\begin{figure}[t]
    \centering
  \includegraphics[width=\columnwidth, height=0.45\columnwidth]{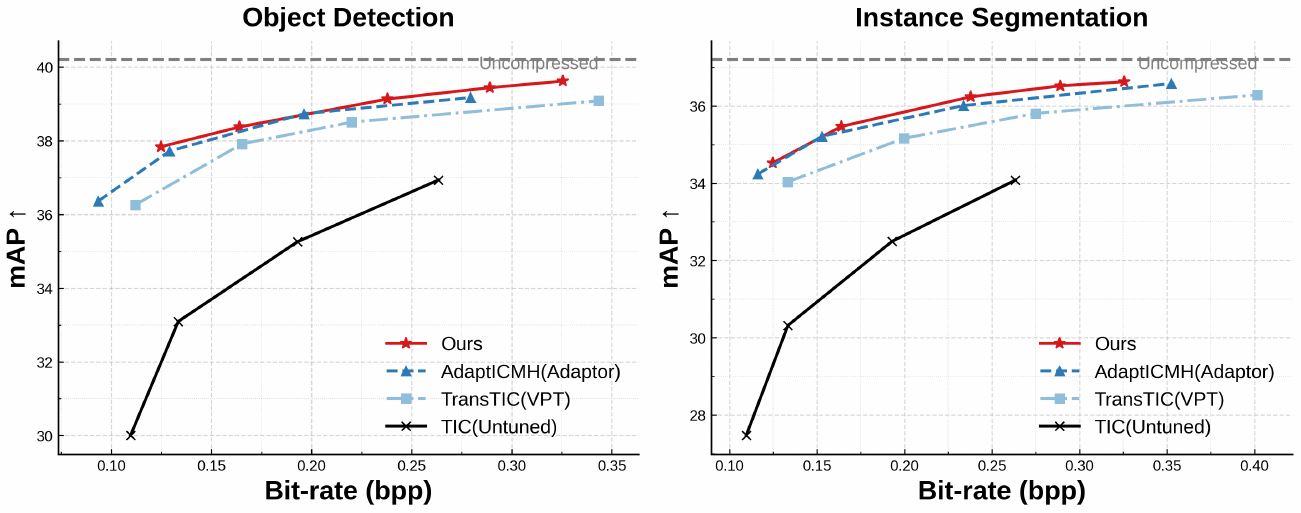}
  \caption{Rate–Accuracy performance comparison of \emph{Object Detection} and \emph{Instance Segmentation} on COCO-2017.
}
  \label{fig:coco}
\end{figure}

To this end, we extend our evaluation to sparse prediction tasks using the COCO 2017~\cite{lin2015microsoftcococommonobjects}, focusing on two representative tasks: \emph{object detection} and \emph{instance segmentation}.
As shown in~\Cref{fig:coco}, our method achieves performance that is comparable to—or even exceeds—that of single-task baselines, demonstrating its robustness and generalizability beyond the dense prediction setting.

\section{Conclusion}
\label{sec:coclu}
\paragraph{\textbf{Conclusion}}
In conclusion, we propose a novel multi-task adaptation framework that enables efficient transfer of a human-centric base codec to a multi-task vision codec. Specifically, we introduce an asymmetric adaptor tuning architecture, which consists of a task-agnostic, single-path encoder adaptor and task-specific, parallel-path decoder adaptors. On top of this structure, we design two feature propagation modules to facilitate both inter-task and inter-scale information exchange. We validate the effectiveness of our framework on multiple benchmark datasets.

Overall, this work presents a practical and effective solution aligned with the coding-for-machine paradigm, and to the best of our knowledge, constitutes the first attempt to enable multi-task transfer in the context of codec adaptation.

\paragraph{\textbf{Limitation}}
As existing work on multi-task codec transfer remains scarce, we primarily compare our method against state-of-the-art single-task optimization baselines.
While optimizing for a single task is inherently simpler than multi-task learning, and we conduct comprehensive evaluations of our approach under single-task, full multi-task, and partial task combination scenarios. Across all settings, the results consistently validate the effectiveness of our proposed multi-task learning framework.
We hope this work provides useful insights and encourages further research in the direction of multi-task codec transfer.

{
    \small
    \bibliographystyle{ieeenat_fullname}
    \bibliography{main}
}

\end{document}